\documentclass{ifacconf}
\usepackage{placeins}
\usepackage{natbib}
\usepackage{graphicx}
\usepackage{tabularx,booktabs}
\usepackage{color}
\usepackage{multicol}
\usepackage{amsmath,amssymb}
\usepackage{amsfonts}
\usepackage{textcomp}
\usepackage{psfrag}
\usepackage{multimedia}
\usepackage{fancybox}

\usepackage{algorithm}
\usepackage{algpseudocode}
\usepackage{amsmath}

\usepackage{varioref}
\usepackage{import}
\usepackage{framed,xcolor}
\usepackage{color}
\usepackage{comment}
\usepackage{graphicx}
\usepackage{subcaption}
\usepackage{float}
\usepackage{etoolbox}
\usepackage{soul}
\usepackage{epstopdf}
\usepackage{algpseudocode}

\definecolor{hos}{rgb}{0.91,0.84,0.42}

\newtheorem{Remark}{Remark}

\begin{document}
\begin{frontmatter}

\title{An Adaptive Coverage Control Approach for Multiple Autonomous Off-road Vehicles in Dynamic Agricultural Fields}

\thanks[footnoteinfo]{This work is supported by the Data Science for Food and Agricultural Systems (DSFAS) program, award no. 2020-67021-40004, from the U.S. Department of Agriculture’s National Institute of Food and Agriculture (NIFA).}

\author[First]{Sajad Ahmadi} 
\author[Second]{Mohammadreza Davoodi}
\author[First]{Javad Mohammadpour Velni}

\address[First]{Dept. of Mechanical Engineering, Clemson University, Clemson, SC 29634 USA (e-mail: \{sahmadi, javadm\}@clemson.edu).}
\address[Second]{Dept. of Electrical and Computer Engineering, The University of Memphis, Memphis, TN 38152 USA (e-mail: mdavoodi@memphis.edu).}

\begin{abstract}                % Abstract of not more than 250 words.
This paper presents an adaptive coverage control method for a fleet of off-road and Unmanned Ground Vehicles (UGVs) operating in dynamic (time-varying) agricultural environments. Traditional coverage control approaches often assume static conditions, making them unsuitable for real-world farming scenarios where obstacles, such as moving machinery and uneven terrains, create continuous challenges. To address this, we propose a real-time path planning framework that integrates Unmanned Aerial Vehicles (UAVs) for obstacle detection and terrain assessment, allowing UGVs to dynamically adjust their coverage paths. The environment is modeled as a weighted directed graph, where the edge weights are continuously updated based on the UAV observations to reflect obstacle motion and terrain variations. The proposed approach incorporates Voronoi-based partitioning, adaptive edge weight assignment, and cost-based path optimization to enhance navigation efficiency. Simulation results demonstrate the effectiveness of the proposed method in improving path planning, reducing traversal costs, and maintaining robust coverage in the presence of dynamic obstacles and muddy terrains.

\begin{comment}
    % This paper addresses the challenge of autonomous field navigation in dynamic agricultural environments. First, an unmanned aerial vehicle (UAV) captures high-resolution images of the field to detect plant positions. Then, the UAV provides aerial imagery for obstacle detection and terrain assessment. The environment, modeled as a weighted directed graph, dynamically updates edge weights based on real-time conditions. Unmanned ground vehicles (UGVs) utilize this graph and employ Dijkstra’s algorithm and a cost function to adjust their paths accordingly in response to weight changes. To ensure a smoother path, a penalty is applied for sharp turns in cost. Finally, simulation results demonstrate the method’s effectiveness in handling dynamic obstacles, minimizing traversal costs, and enhancing adaptability.

% In this work, we address the distributed coverage control problem for deploying a team of heterogeneous robots with nonlinear dynamics in a partially known environment represented as a weighted mixed graph. The environment includes dynamic obstacles, such as muddy soil patches, which move across edges in the graph. These obstacles are assigned higher weights, and their influence propagates to neighboring edges based on their trajectory. To optimize coverage, the robot trajectories are adaptively trimmed by predicting and accounting for the obstacle's movement. This approach ensures efficient and safe navigation while maintaining robust coverage in dynamic and uncertain environments.
\end{comment}

\end{abstract}


\end{frontmatter}
%===============================================================================

\section{Introduction}
\vspace{-2mm}
Growing population has driven a rising demand for food, putting significant pressure on crop and livestock production. This strain often leads to environmental concerns and shortages of trained agricultural labor \citep{bechar2016agricultural}. In response, smart farming technologies have gained prominence and offer solutions to enhance productivity while minimizing waste and operational expenses \citep{singh2021agrifusion}. At the forefront of this transformation is precision agriculture (PA), which leverages advanced data analytics, autonomous systems like UAVs and UGVs, and controls/automation to optimize field operations \citep{monteiro2021precision}.

In PA, the combination of UAVs and UGVs can make a significant impact \citep{mammarella2022cooperation}. UAVs are capable of detecting areas that require attention, while UGVs can incorporate the data received from UAVs into their path planning for more efficient navigation \citep{bechar2016agricultural,  munasinghe2024comprehensive}. Meanwhile, agricultural fields are generally subject to continuous changes, with obstacles such as moving machinery, livestock, and terrain variations - such as muddy soil patches - introducing uncertainty into path planning and coverage control \citep{etezadi2024comprehensive}. Traditional coverage control algorithms typically assume static environments, limiting their applicability in real-world farming conditions \citep{schwager2009gradient}. Similarly, existing path planning approaches such as Dijkstra's algorithm, A*, and Hybrid A* also rely on predefined maps and static assumptions \citep{dolgov2008practical}. Therefore, these methods do not adapt to moving obstacles or terrain variations in real-time. More recent approaches in multi-agent path finding (MAPF) have attempted to address dynamic environments by incorporating collision avoidance mechanisms between agents \citep{stern2019multi}.

An algorithmic solution for persistent coverage has been proposed, where robots use fast marching methods and a coverage action controller to maintain the desired coverage level efficiently and safely \citep{palacios2017optimal}. Also, a coordination strategy for a hybrid UGV–UAV system in planetary exploration has been presented, in which the UGV serves as a moving charging station for the UAV to optimize target point coverage while minimizing travel distance \citep{ropero2019terra}. The authors introduce a terrain-aware path planning method for UGVs based on the Hybrid A* algorithm, optimizing both traversability and distance to improve autonomous navigation in rough terrain \citep{thoresen2021path}. A prioritized path-planning algorithm for multi-UGV systems in agricultural environments extends MAPF by incorporating robot priorities to reduce congestion without inter-robot communication \citep{jo2024field}. A multi-phase approach for cooperative UAV–UGV operations in precision agriculture focuses on automated navigation and task execution in complex, unstructured environments such as sloped vineyards \citep{mammarella2020cooperative}. A partitioning algorithm and deployment strategy have been developed for distributing heterogeneous autonomous robots in a partially known environment, optimizing coverage and resources for applications like agricultural field monitoring \citep{davoodi2020heterogeneity}. Finally, a new partitioning algorithm based on a state-dependent proximity metric and a discounted cost function for robots with nonlinear dynamics has also been proposed \citep{davoodi2021graph}.

While the aforementioned methods provide effective partitioning and tracking strategies, they often fail to dynamically account for moving obstacles or changing terrain conditions in a graph. \textbf{\textit{The contribution and novelty of this study lies in explicitly incorporating obstacle avoidance and adaptation to terrain conditions in path planning. To this end, we propose an adaptive coverage control strategy that integrates UAV-based observations with UGV path planning. Specifically, once the UAV detects an obstacle or obtains data on the terrain conditions, the path planning for UGVs adjusts coverage paths accordingly to ensure a safe and efficient navigation}}.

The remainder of this paper is structured as follows. Section \ref{sec:2} defines the problem and formalizes the environment modeling. Section \ref{sec:3} introduces our adaptive graph-based coverage control strategy, including Voronoi partitioning and dynamic path updating. Section \ref{sec:4} presents simulation results, and Section \ref{sec:5} gives concluding remarks.

% The remainder of this paper is organized as follows. Section \ref{sec:2} presents the problem formulation and our proposed methodology. Section \ref{sec:3} discusses simulation results and evaluates system performance. Finally, Section \ref{sec:4} concludes the paper and outlines potential directions for future research.

\vspace{-1mm}
\section{Problem Statement and Preliminaries} \label{sec:2}
\vspace{-2mm}
%The primary objective of this paper is to address the autonomous monitoring of a partially known environment, denoted as  $Q$, where there can be dynamic obstacles or terrain conditions in the environment(which is the UGVs' workspace).
The primary objective of this paper is to address the problem of autonomous monitoring of an agricultural field, represented as a partially known environment $Q$, using a group of UGVs. This environment may contain both dynamic obstacles (e.g., moving vehicles) and static obstacles (e.g., muddy regions). This section explains the process of modeling the agricultural field as a weighted directed graph and defines the problem.
\begin{comment}
    
\subsection{Modeling of unmanned ground vehicle} \label{sec:model}

In this study, the UGVs are considered to be non-holonomic robots. The kinematic model of the $k$th robot, for $k \in \mathcal{K} = \{1, \ldots, n\}$ is described by the following dynamic equations
\begin{align}
    \begin{bmatrix}
      \dot{x}^k\\
       \dot{y}^k\\
       \dot{v}^k\\
       \dot{\theta}^k
     \end{bmatrix}
    = 
      \begin{bmatrix}
   v^k\cos{\theta^k} \\
   v^k\sin{\theta^k} \\
   a^k \\
   \omega^k
   \end{bmatrix},
     \label{eq:model}
\end{align}
where $x^k$ and $y^k$ are the Cartesian coordinates of the system in world reference frame, $\theta^k$ denotes its orientation, and $v^k$ and $a^k$ corresponds to the velocity and acceleration, respectively.

\end{comment}
\vspace{-2mm}
\subsection{Environment Modeling}
\vspace{-2mm}
The environment, representing the robots' workspace, is modeled as a weighted directed graph $\mathcal{G}(\mathcal{V}, \mathcal{E}, \mathcal{C})$ that consists of a node set $\mathcal{V} = \{v_1,v_2, \ldots, v_m\}$, an edge set $\mathcal{E} \subseteq \mathcal{V} \times \mathcal{V}$, and associated weights $\mathcal{C} \in \mathbb{R}^{m \times m}$. The set of neighboring nodes for a given node $x$ in the graph is denoted by $\mathcal{N}_{\mathcal{G}}(x) =\{y \in \mathcal{V} \mid \overrightarrow{x y} \in \mathcal{E}\}$.

\begin{figure}[h!]
    \centering
    % First subfigure
    \begin{subfigure}[t]{0.32\linewidth} % Adjust width as needed
        \centering
        \includegraphics[width=\linewidth]{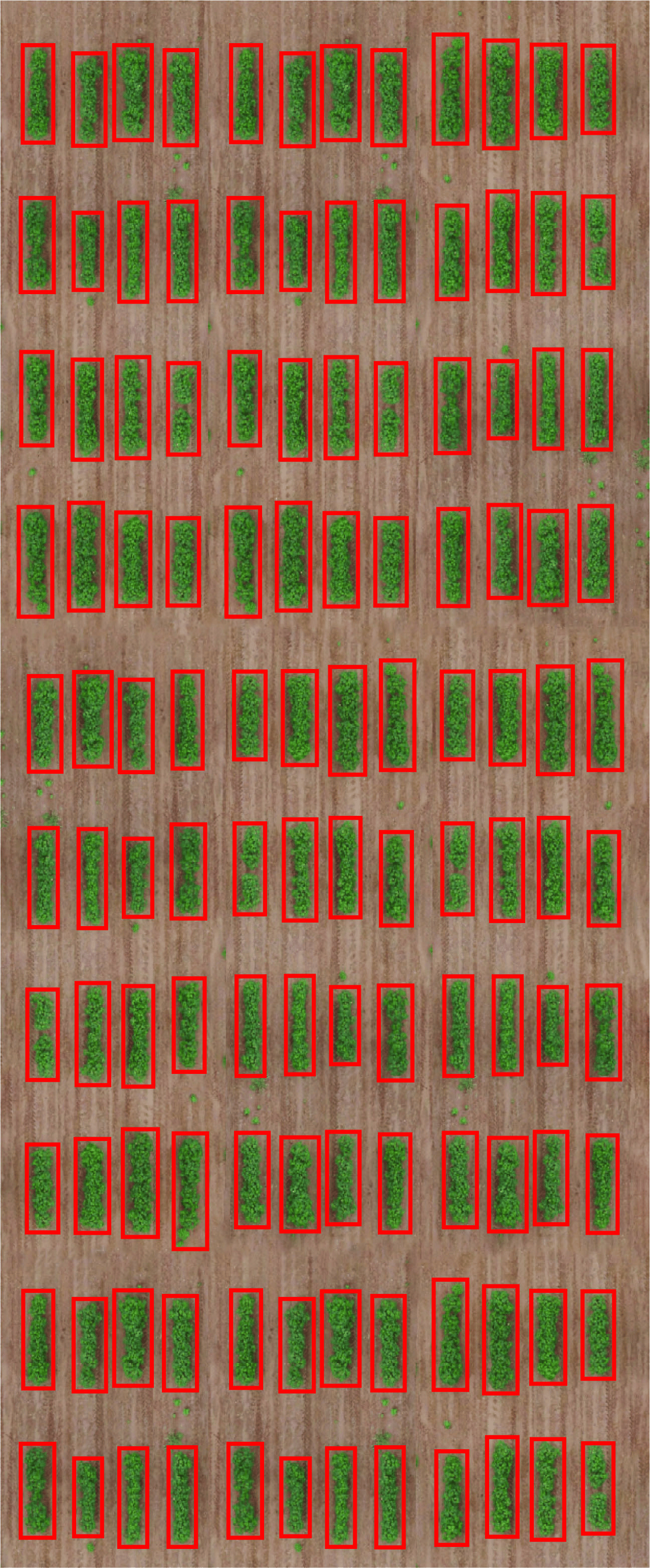} % Replace with your file
        \caption{}
        \label{fig:detection_a}
    \end{subfigure}
    % Add horizontal space
    % \hspace{0.05\linewidth}
    % Second subfigure
    \begin{subfigure}[t]{0.32\linewidth} % Adjust width as needed
        \centering
        \includegraphics[width=\linewidth]{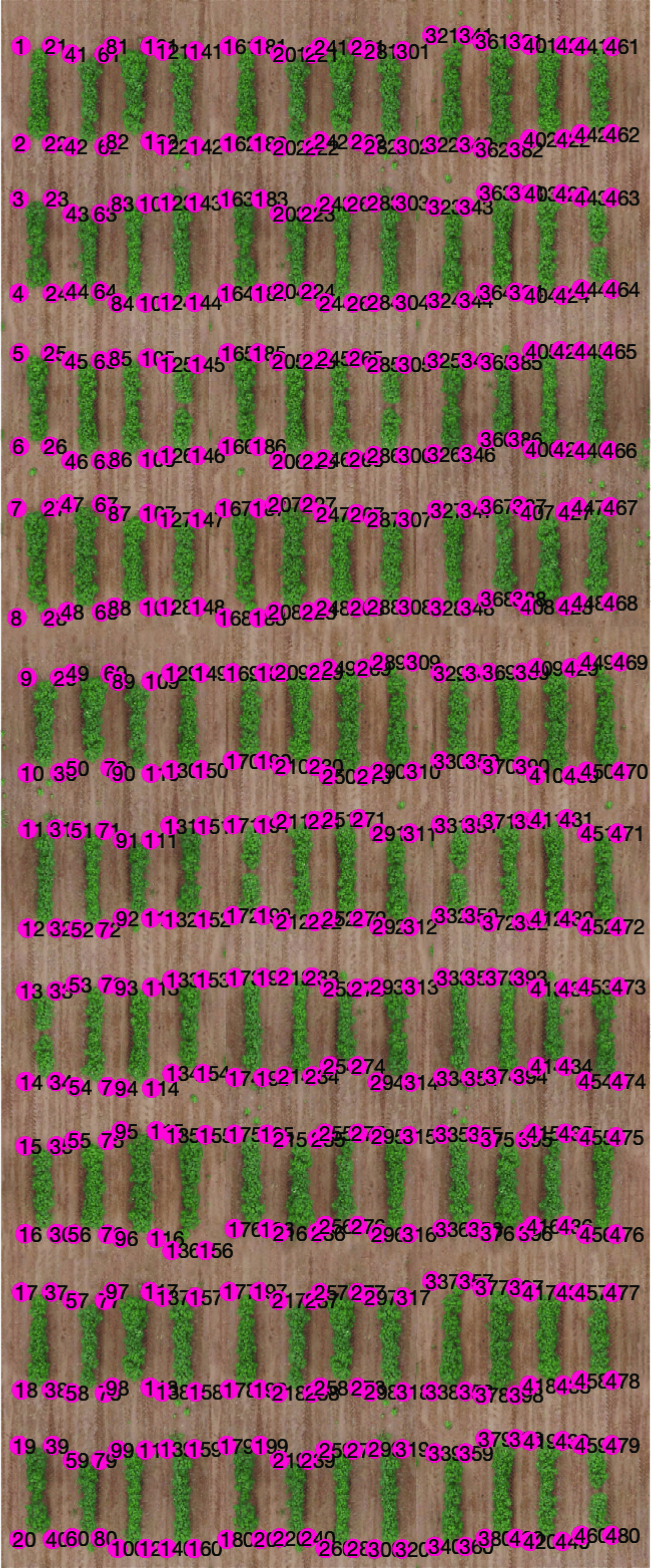} % Replace with your file
        \caption{}
        \label{fig:detection_b}
    \end{subfigure}
    % Add horizontal space
    % \hspace{0.05\linewidth}
    % Third subfigure
    \begin{subfigure}[t]{0.32\linewidth} % Adjust width as needed
        \centering
        \includegraphics[width=\linewidth]{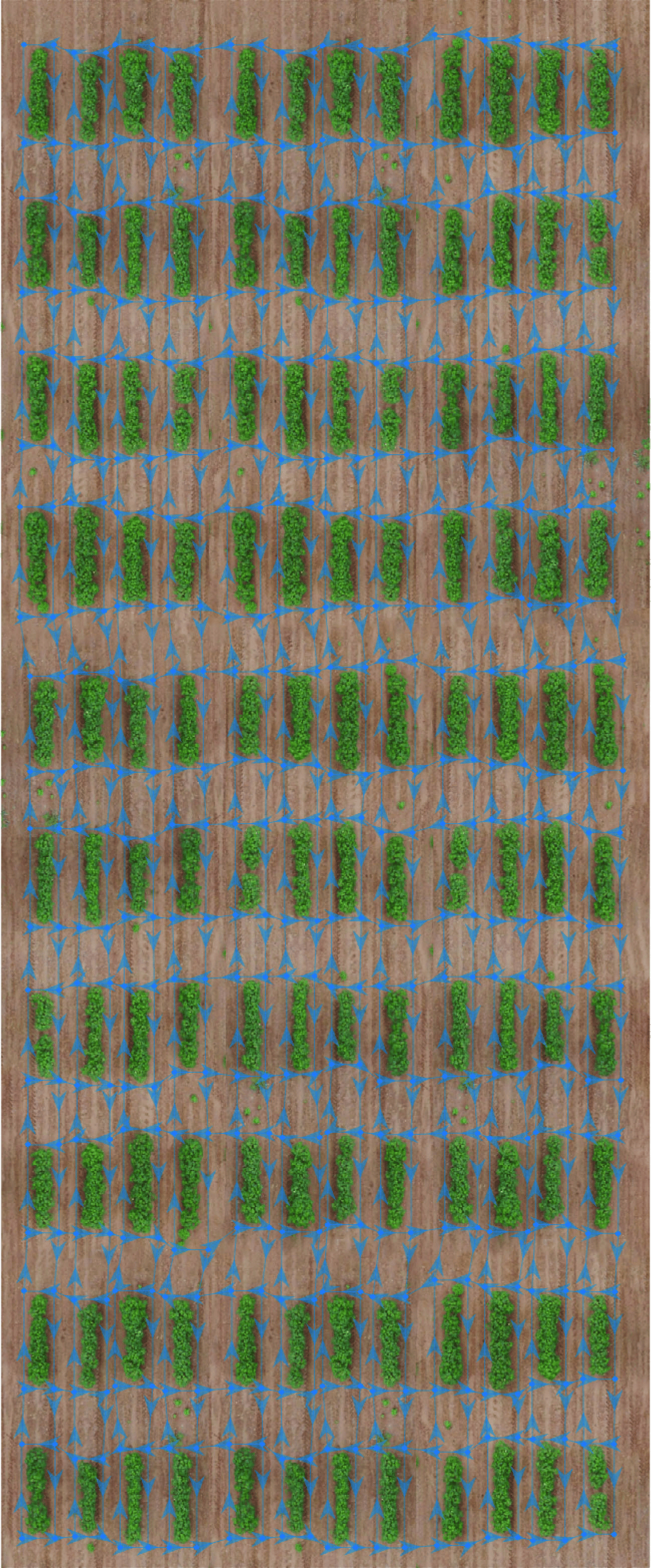} % Replace with your file
        \caption{}
        \label{fig:detection_c}
    \end{subfigure}   
    \caption{Illustration of plant detection and graph construction: (a) detected plant rows represented as bounding rectangles, (b) graph nodes generated at the corners of the rectangles, and (c) directed edges added to form a complete
graph structure.}
    \label{fig:detection}
\end{figure}

Assuming that an image of the agricultural field is provided by the UAV, plant rows are represented by bounding boxes (rectangles) using image detection (see Fig. \ref{fig:detection}(a)). Each corner of these rectangles is used to define nodes for the graph, indicated by red circles in Fig. \ref{fig:detection}(b). Node numbering begins from the top left of the field, moving row by row, then column by column. These nodes are connected to each other with directed straight lines, as shown in Fig. \ref{fig:detection}(c), to construct the graph. In this study, nodes are viewed as specific points within the environment, acting as predefined destinations for robots departing from their current positions, while edges represent the paths for robot movement between nodes. Although robot movement is restricted to straight lines along the edges, this approach can also be extended to more complex movements. Furthermore, the weights associated with these edges are described in Section \ref{sec:edge_update}.

\begin{comment}
 To construct the graph, first we use the image of the field captured by drone and detect plant rows and show them as a bounding rectangles(Fig. \ref{}). Then we define each corner of the rectangle as a node for our graph(Fig. \ref{}). Node numbers are started from top left side of the field and for each rows and then next column in the field. Then, we use straight lines to connect each nodes so the edges of the graph is created.
To construct the graph, we investigate the nodes as specific points within the environment. These nodes correspond to predefined locations that robots must reach after departing from their current positions (see figure \ref{fig:detection}). In this study, we focus on scenarios where robot movement between nodes is restricted to straight-line paths. Consequently, the edges in the graph are assumed to be straight-line connections. However, it is possible to extend this approach to applications with more complex movement, without compromising generality. The weights associated with these edges are detailed in Section \ref{sec:edge_update}.    
\end{comment}

Once the field is represented as a graph, a density function $\phi(v): \mathcal{V} \rightarrow \mathbb{R}^{+}$ is introduced over $\mathcal{G}$ to highlight regions of interest, meaning nodes with higher priority for servicing. In agricultural applications, these regions of interest may correspond to areas containing plants affected by biotic or abiotic stresses or exhibiting specific phenotypic traits, such as flowering or water accumulation in the crop field, among others. The density function $\phi(v)$ is derived from a continuous function defined over the original environment. Essentially, $\phi(v)$ is assigned larger values for nodes near the center, while nodes farther from the center receive smaller values.
\vspace{-2mm}
\subsection{Problem Formulation}
\vspace{-2mm}
Consider a team of $n$ UGVs, denoted as $r^k$, $k \in \mathcal{K} = \{1, \ldots, n\}$, with initial positions given by $v_i^k = \left(x_i^k, y_i^k\right) \in \mathcal{V}$. Also, Each UGV is equipped with the necessary sensors, cameras, or actuators to perform its tasks.

Assumption 1: All UGVs have access to the graph $\mathcal{G}$ and possess complete knowledge of the density function $\phi: \mathcal{V} \rightarrow \mathbb{R}^{+}$.

\textbf{Problem:} \textit{Develop a graph-based distributed coverage control strategy to deploy a team of UGVs for monitoring critical regions within the environment $Q$ while avoiding muddy patches or dynamic obstacles.}

\textbf{Practical value of the work:} Let us consider a fleet of autonomous UGVs equipped with a targeted-spray boom for pesticide application. A UAV first surveys the field (from above the canopy) to identify pest hotspots and communicates obstacle and terrain data (e.g., moving harvesters, irrigation machinery, muddy patches) to the UGVs. An adaptive coverage algorithm should allow the UGVs to dynamically adjust their path—avoiding obstacles in real time, ensuring complete, efficient coverage of the affected zones while reducing chemical use and labor. This scenario exemplifies just one of many practical applications that can benefit from our proposed methodology.

\vspace{-2mm}

\section{Methodology}\label{sec:3}
\vspace{-2mm}
%In this section, we formalize the weight assignment and partitioning algorithm, as well as a cost-based path optimization strategy to minimize traversal costs. 
The main results of this work are provided in this section.
The subsequent subsections provide detailed descriptions of the edge weight assignment, partitioning strategy, and optimization framework that enable robust and adaptive coverage control. Also, the control strategy for an individual UGV and reference trajectory generation are adopted from \citep{davoodi2020heterogeneity}.
% This section formalizes the weight assignment process for the edges of a graph representation of the environment and describes the partitioning of the workspace using Voronoi-like cells. The control strategy for an individual UGV and reference trajectory generation are adopted from \citep{davoodi2020heterogeneity}.

% In this part, we explain how we assign weight to the graph edges, and also we discuss regarding partitioning of the environment using Voronoi-like cells. Controlling of a single robot(UGV) and reference gerating are sourced from \citep{davoodi2020heterogeneity}.
\vspace{-1mm}
\subsection{Edge Weight Assignment in the Graph} \label{sec:edge_update}
\vspace{-1mm}
Consider the  graph \( \mathcal{G}(\mathcal{V}, \mathcal{E}, \mathcal{C}) \). Initially, each edge $e_{ij} \in \mathcal{E}$ (connecting node $v_i$ to node $v_j$) is assigned a weight of $c_{ij}$ that is the Euclidean distance between the $v_i$ and $v_j$ in the agricultural field (see Fig. \ref{fig:detection}). Also, it is assumed that the UAV is capable of perceiving obstacle positions, velocities, and terrain conditions. 

If an obstacle is detected on the edge connecting nodes $v_m$ and $v_n$ (i.e., $e_{mn}$), the weight corresponding to that edge is increased to reflect the obstacle’s impact as
\begin{equation}
\label{eq:obstacle}
    c_{mn} \leftarrow c_{mn} + \alpha .\exp({v_{\text{obs}}^2}/{v^2_{0}}),
\end{equation}
where $\alpha $ and $v_{0}$ are scaling factors modulating the influence of the obstacle, and $v_{\text{obs}}$ denotes the velocity of the detected obstacle.

Furthermore, to discourage paths that intersect with the predicted obstacle trajectory, the weights of edges along the obstacle’s motion direction are penalized. The affected edges are selected based on the obstacle’s velocity and graph connectivity, ensuring they align with the predicted path. Specifically, if an obstacle moves along an edge $e_{mn}$, the weights of $N$ subsequent edges in its motion direction are modified as
\begin{equation} 
\label{eq:sub_obstacle}
    c_{ij} \leftarrow c_{ij} + \alpha \exp({v_{\text{obs}}^2}/{v^2_{0}}) \cdot \exp(-{d_{edge}(e_{ij}, e_{mn})}/{d_0}),
\end{equation}

where $ c_{ij}$ is the weight of the $ij$-th affected edge, $d_{edge}(e_{ij}, e_{mn})$ is computed as the sum of edge weights along the obstacle path connecting them, including $e_{mn}$ and $e_{ij}$ themselves. And, $d_0$ is a scaling factor that controls the decay rate of the penalty with distance.

Furthermore, to account for variations in terrain conditions, such as muddy areas, the weight of each edge of those areas is adjusted as
% corresponding to row \( i \) and column \( j \) in \( \mathcal{C} \) is adjusted as:
\begin{equation}
\label{eq:muddy}
    c_{mn} \leftarrow c_{mn} + \beta . \exp(T_{mn}),
\end{equation}
where $\beta$ is a scaling factor, and $T_{mn}$ represents the terrain condition of the edge. The UAV will assess the condition and provide $T_{mn}$.

\begin{Remark}
    The graph \( \mathcal{G} \) is dynamically updated as obstacles move or terrain conditions change, ensuring that the edge weights reflect the latest environmental changes.
\end{Remark}
\vspace{-1mm}
\subsection{Graph Partitioning and Cost Function} 
\vspace{-1mm}
\label{sec:graph_partitioning}
After updating the graph $\mathcal{G} = (\mathcal{V}, \mathcal{E}, \mathcal{C})$, consider a set of $n$ UGVs positioned at designated nodes $\{v_1, v_2, \dots, v_n\} \subset \mathcal{V}$. The goal is to partition $\mathcal{G}$ into $k$ disjoint subgraphs, denoted as $\{\mathcal{V}_1, \mathcal{V}_2, \dots, \mathcal{V}_n\}$, where each subgraph $\mathcal{V}_k$ represents a Voronoi-like cell assigned to agent $k$.  

Next, we define a quantity $J^k_{v_i^k , v_j}(P^k)$ for each robot $r_k$ going from node $v^k_i$ to $v_j$ through path $P^k$. This function is designed to favor paths that minimize both distance and turning effort, ensuring a cost efficient movement. Formally, it is expressed as
\begin{equation}
    J^k_{v_i^k , v_j} = \min_{P^k \in \mathcal{P}_{k}} \sum_{e_{ab} \in P^k} \Big(c_{ab} + \lambda C_{\text{turn}}(P^k)\Big)
    \label{eq:Jij}
\end{equation}
where $\mathcal{P}_k$ is the set of all the paths between the robot position node $v_i^k$ and node $v_j$ for robot $k$. The term $c_{ab}$ represents the weight associated with edge $e_{ab}$, and the summation accumulates these weights over all edges in the selected path. Additionally, $C_{\text{turn}}(P^k)$ is defined as
\begin{equation}
    C_{\text{turn}}(P^k) = \sum_{j=1}^{k-1} \mathbb{I}(\theta_j \geq 90^\circ),
\end{equation}
where \( \mathbb{I}(\theta_j \geq 90^\circ) \) is an indicator function that returns 1 if the turning angle $(\theta_j)$ between consecutive edges \( e_{ij} \) and \( e_{jk} \) is greater than or equal to 90 degrees, and 0 otherwise. The weighting parameter \( \lambda \) in (\ref{eq:Jij}) allows for a trade-off between path length and smoothness, effectively controlling the preference for minimizing turns while maintaining efficiency in path selection.

Building upon this framework, as well as the results from \citep{yun2014distributed}, the Voronoi-like partitions $\mathcal{V}_k$ generated by robot $k$ are defined as
\begin{equation}
\label{eq:voronoi}
    \mathcal{V}_k = \left\{\mathcal{V}_k \in \mathcal{V} | J_{v_i^k , v_j^k} < J_{v_{i}^{k^{'}} , v_{j}^{k^{'}}}  \right\},
\end{equation}
where $k^{'}$ represents all robots in $\mathcal{K}$ except for the robot $k$. Note that the robot $k$ is responsible for monitoring all the events occurring within its assigned region $\mathcal{V}_k$. Then, the general deployment problem is formulated as minimizing the following cost function
\begin{equation}
\label{eq:cost_function}
    \mathcal{H}(v_i, \mathcal{G})=\sum_{k=1}^n \sum_{v_j \in \mathcal{V}_i} J^k_{v_i^k, v_j} \tilde{\phi}\left(v_i^k, v_j\right),
\end{equation}
where
\begin{equation}
\label{eq:density}
    \tilde{\phi}\left(v_i^k, v_j\right)= \begin{cases}\phi(v_j) & \text { if } v_i^k \neq v_j, \\ 0 & \text { if } v_i^k=v_j.\end{cases}
\end{equation}

It is noted that the function $\tilde{\phi}\left(v_i^k, v_j\right)$ prevents repetitive monitoring of the regions of interest by reducing their importance to zero after they are visited by one of the robots. 

Next, an iterative approach is proposed to navigate the robots between the nodes, ensuring a continuous reduction in the locational optimization cost, $\mathcal{H}$, until all nodes with a nonzero value in $\phi$ have been visited. After each movement, the graph edge weights ($\mathcal{C}$) and partitions are updated. Subsequently, for each robot, the cost at its current position is evaluated and compared with the costs at its neighboring nodes. The robot moves to the neighboring node with the lowest cost if it offers a reduction in $\mathcal{H}$. If no neighboring node provides a lower cost, the robot remains in place until changes in the environment—such as the removal of an obstacle—allow for further movement. Also, if a robot reaches a node with a nonzero value in $\phi$, the value of $\phi$ at that node is set to zero.

The detailed methodology is presented in Algorithm \ref{alg:template}.

\begin{algorithm}
    \caption{Path Planning for UGVs}
    \label{alg:template}
    \begin{algorithmic}[1]
        \State \textbf{Inputs:}
        \Statex \hspace{1em} 1. $\mathcal{G}(\mathcal{V}, \mathcal{E}, \mathcal{C})$
        \Statex \hspace{1em} 2. $\{v_i^1, \dots, v_i^n\}$
        \Statex \hspace{1em} 3. $\phi$
        \Statex \hspace{1em} 4. $\alpha, \beta, v_{obs}, v_0, d_0, \lambda, N$
        
        \State \textbf{Output:} Compute next best point for each robot.
        
        \State $iter \gets 1$
        \While{$\phi \neq zeros(size(\phi))$}
            \State Update the UGV’s location, obstacles’ positions and velocities, and directions.
            \State Update edges' weights (Eqs.~\eqref{eq:obstacle}--\eqref{eq:muddy}).
            \State Update Voronoi-like partitioning (Eq.~\eqref{eq:voronoi}). 
            \State $\mathcal{H}^{iter}_k \gets$ compute cost function for current position of each UGV $k$.
            \For{each UGV $k \in \mathcal{K}$}
                \State Compute the cost function for $v_i^k$'s neighbor nodes $\mathcal{N}_{\mathcal{G}}(v_i^k)$.
                \State  $\mathcal{H}_a \gets $ minimum cost between neighboring nodes ($v_i^k$).
                \If{$\mathcal{H}_a < \mathcal{H}^{iter}_k$}
                    \State $v_i^k \gets \text{corresponding node for } \mathcal{H}_a$
                \EndIf
                \If{$\phi(v_i^k) > 0$}
                    \State $\phi(v_i^k) \gets 0$
                \EndIf
            \EndFor
            \State $iter \gets iter + 1$
        \EndWhile
    \end{algorithmic}
\end{algorithm}

\vspace{-3mm}
\section{Simulation Results and Analysis} \label{sec:4}
\vspace{-3mm}
As previously discussed, the proposed coverage control method accounts for dynamic (time varying) and static obstacles such as moving machinery and/or muddy soil patches. This section examines various aspects and capabilities of the developed approach through different case studies. The first case study analyzes a dynamic obstacle in the field and its impact on navigation. It includes a scenario for “regular dynamic obstacle,” where obstacles are detected since entering the field and a “sudden dynamic obstacle” scenario, where they fail to be detected from the beginning. The second case investigates muddy soil patches and their effect on navigation accuracy. The scalars used throughout our simulation studies are $\alpha = 5m, \beta=1m, |v_{obs}|= 3 m/s, v_0 = 1 m/s, \lambda =0.1m, d_0 =1m$, and $N=3$.

% In this section, we present the simulation setup and the evaluation of the proposed coverage control system for autonomous field navigation for UGVs using the data gathered by the UAV. The simulation models various real-world challenges, including plant detection, dynamic obstacle avoidance, terrain variations, and unexpected obstacles. By utilizing a graph-based approach, the system adjusts the UGV’s path in real-time based on input from the drone, considering factors such as muddy areas and slopes. We explore different scenarios, such as the drone detecting and relaying obstacle positions and the system adapting to sudden changes in the environment. The following subsections provide a detailed description of the simulation methodology, results, and performance evaluation.

% \subsection{Field Image Acquisition and Plant Detection}
% In this section, we describe the method used by the drone to acquire images of the field and detect plant positions. The UAV captures high-resolution images and processes them using computer vision techniques  to detect plant positions (Fig. \ref{fig:detection}). We employ a combination of image segmentation and contour detection to identify individual plants. Each detected plant is enclosed within a bounding rectangle, with the corners of these rectangles serving as graph nodes for subsequent path planning.

\vspace{-2mm}
\subsection{UGV Path Planning with Dynamic Obstacle Avoidance}
\begin{figure}[h!]
    \centering
    % First subfigure
    \begin{subfigure}[t]{0.46\linewidth} % Adjust width as needed
        \centering
        \includegraphics[width=\linewidth]{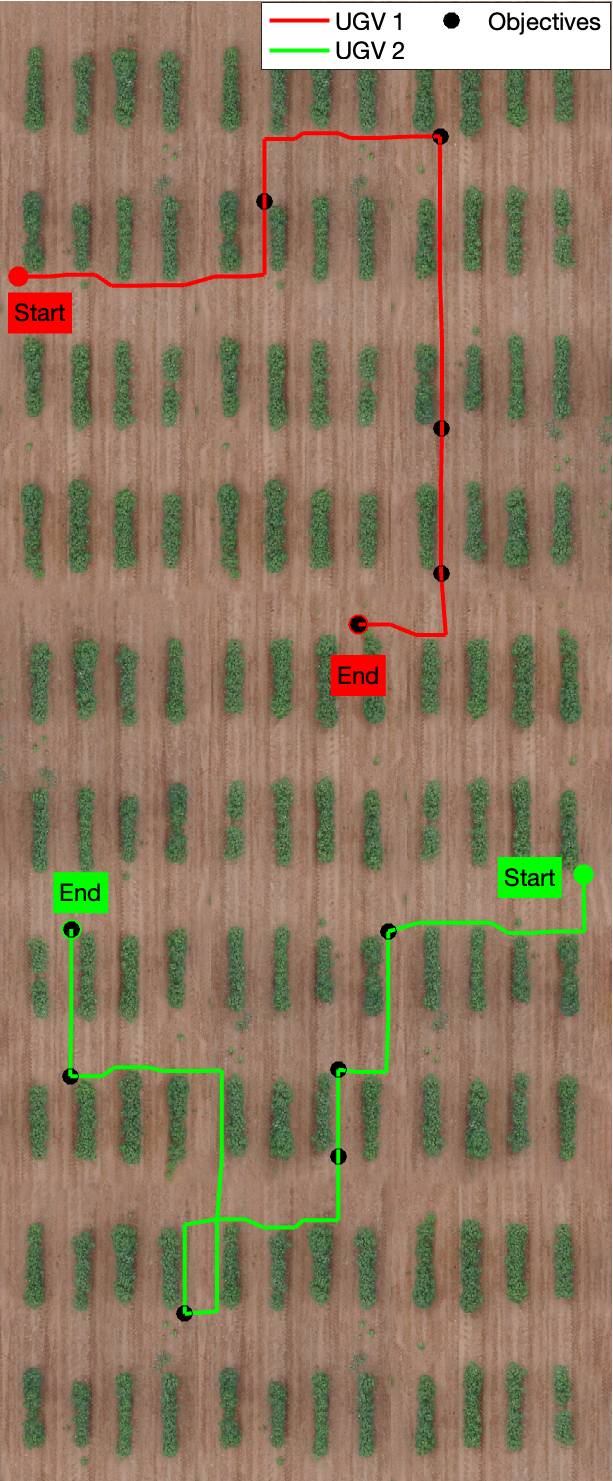} % Replace with your file
        \caption{Without obstacle}
        \label{fig:obs1_a}
    \end{subfigure}
    % Add horizontal space
    % \hspace{0.05\linewidth}
    % Second subfigure
    \begin{subfigure}[t]{0.46\linewidth} % Adjust width as needed
        \centering
        \includegraphics[width=\linewidth]{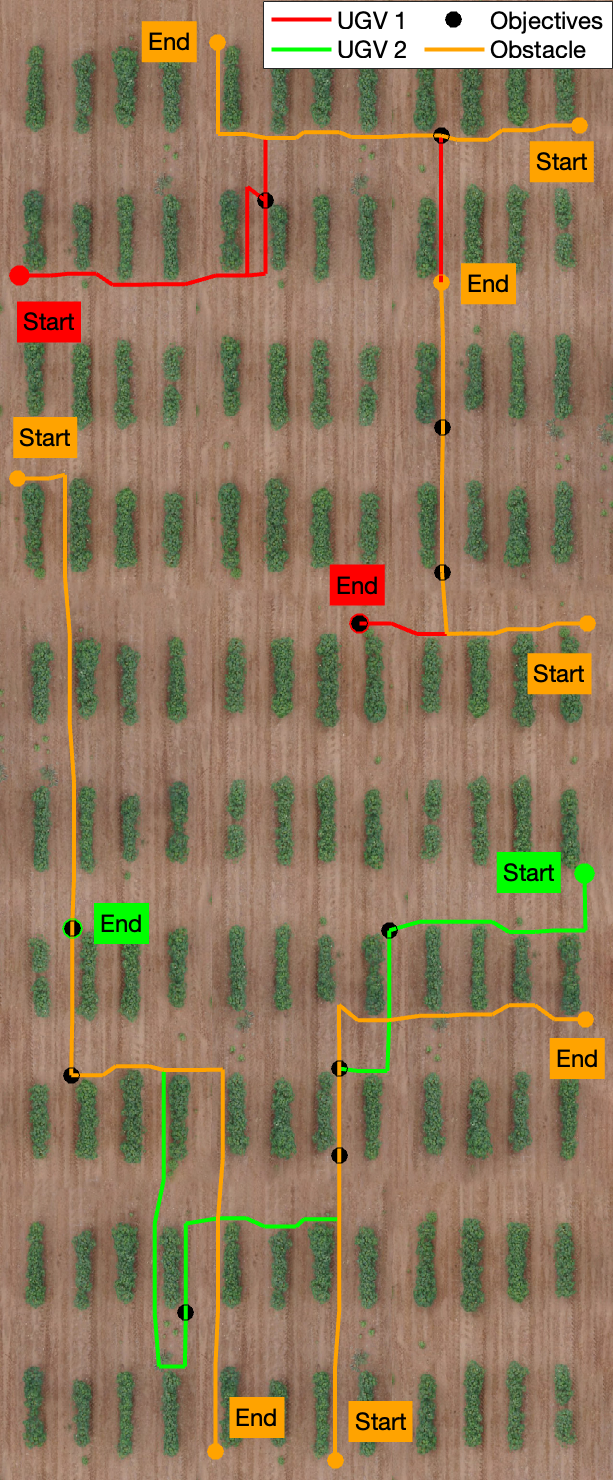} % Replace with your file
        \caption{With obstacle}
        \label{fig:obs1_b}
    \end{subfigure}

    \caption{Comparison of UGV paths with and without the (moving) obstacles, indicating the impact of dynamic obstacle avoidance. See Fig. \ref{fig:regular_3D} for the overlap between the UGV and obstacle trajectories.}
    \label{fig:obs1}
\end{figure}

\begin{figure}[h!]
    \centering
    % First subfigure
    \begin{subfigure}[t]{0.49\linewidth} % Adjust width as needed
        \centering
        \includegraphics[width=\linewidth]{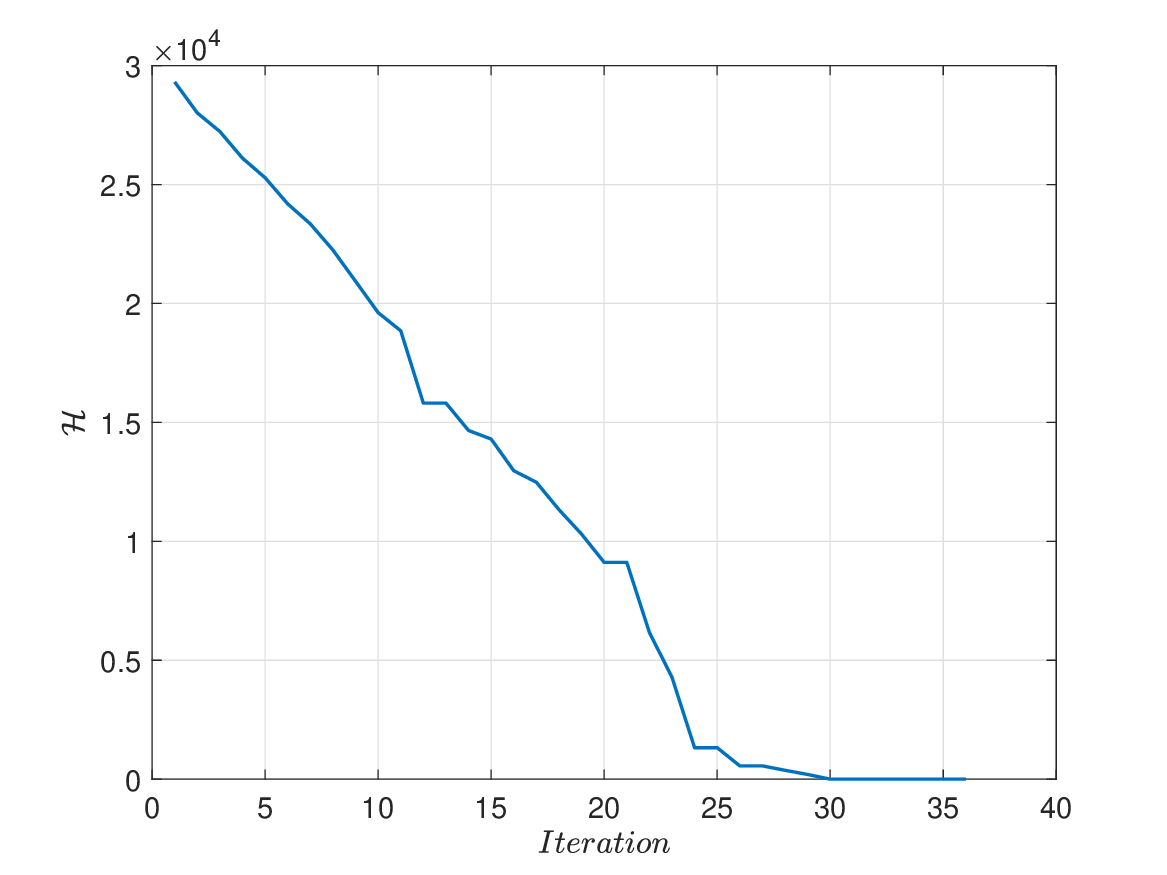} % Replace with your file
        \caption{Without obstacle}
        \label{fig:no_obstacle_cost}
    \end{subfigure}
    % Add horizontal space
    % \hspace{0.05\linewidth}
    % Second subfigure
    \begin{subfigure}[t]{0.49\linewidth} % Adjust width as needed
        \centering
        \includegraphics[width=\linewidth]{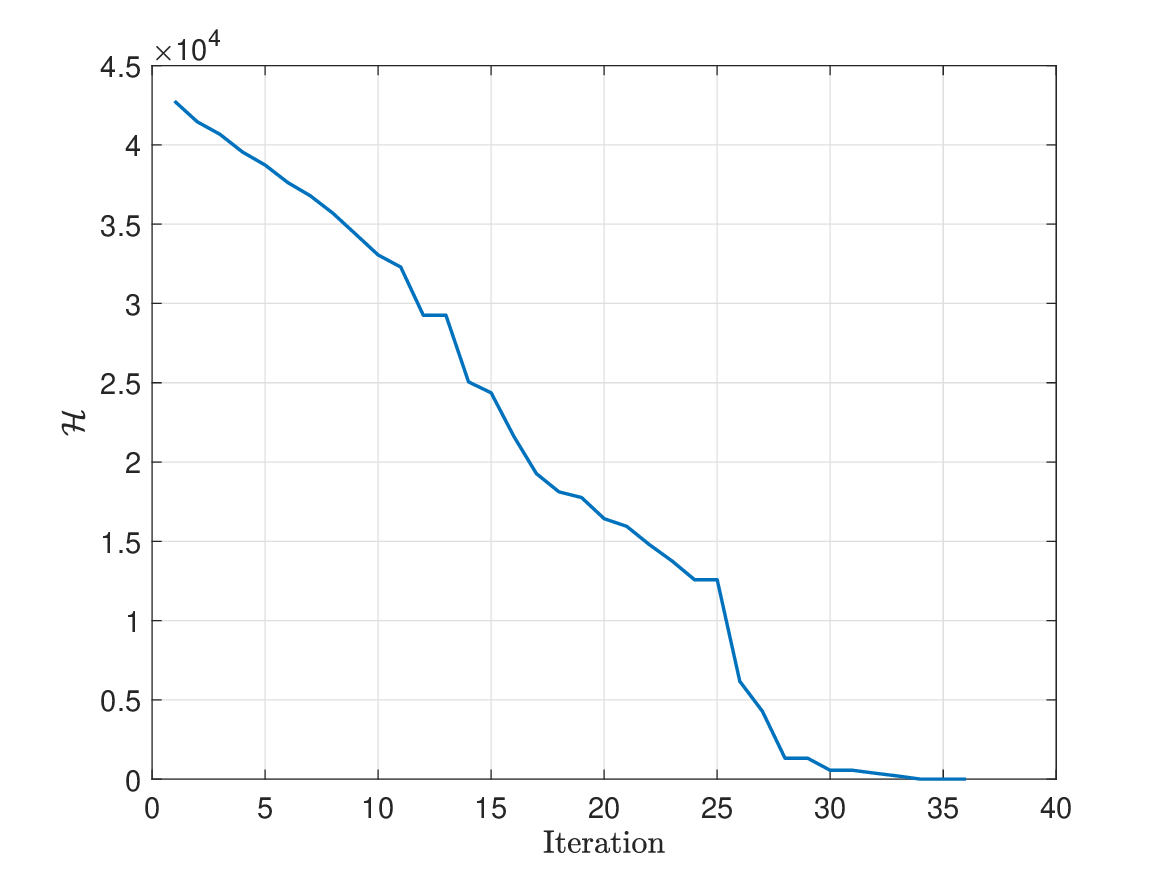} % Replace with your file
        \caption{With obstacle}
        \label{fig:with_obstacle_cost}
    \end{subfigure}

    \caption{Impact of dynamic obstacles on UGVs' coverage cost. Comparing the two plots clearly shows an increase in the cost due to the deviation of the UGVs from their optimal path to avoid dynamic obstacles; this also leads to longer time for the cost to converge to zero.}
    \label{fig:obs1_cost}
\end{figure}

\vspace{-2mm}
The path planning strategy for the UGVs dynamically adapts to environmental changes by leveraging real-time data from the UAV. The UAV continuously tracks moving obstacles, providing their positions and velocities to the UGV, which integrates this information into a graph-based path-planning algorithm. The UGV continuously updates its trajectory to ensure collision-free navigation while maintaining efficiency. This is why we refer to it as adaptive. This adaptive mechanism enables safe and effective operation in dynamic agricultural environments. Figure \ref{fig:obs1_a} shows the UGVs' trajectories when there is no obstacle , while Figure \ref{fig:no_obstacle_cost} shows the coverage cost associated with it. These results serve as a baseline for comparison with other scenarios.  

\begin{figure}[h!]
    \centering
    % First subfigure
    \begin{subfigure}[t]{0.6\linewidth} % Adjust width as needed
        \centering
        \includegraphics[width=\linewidth]{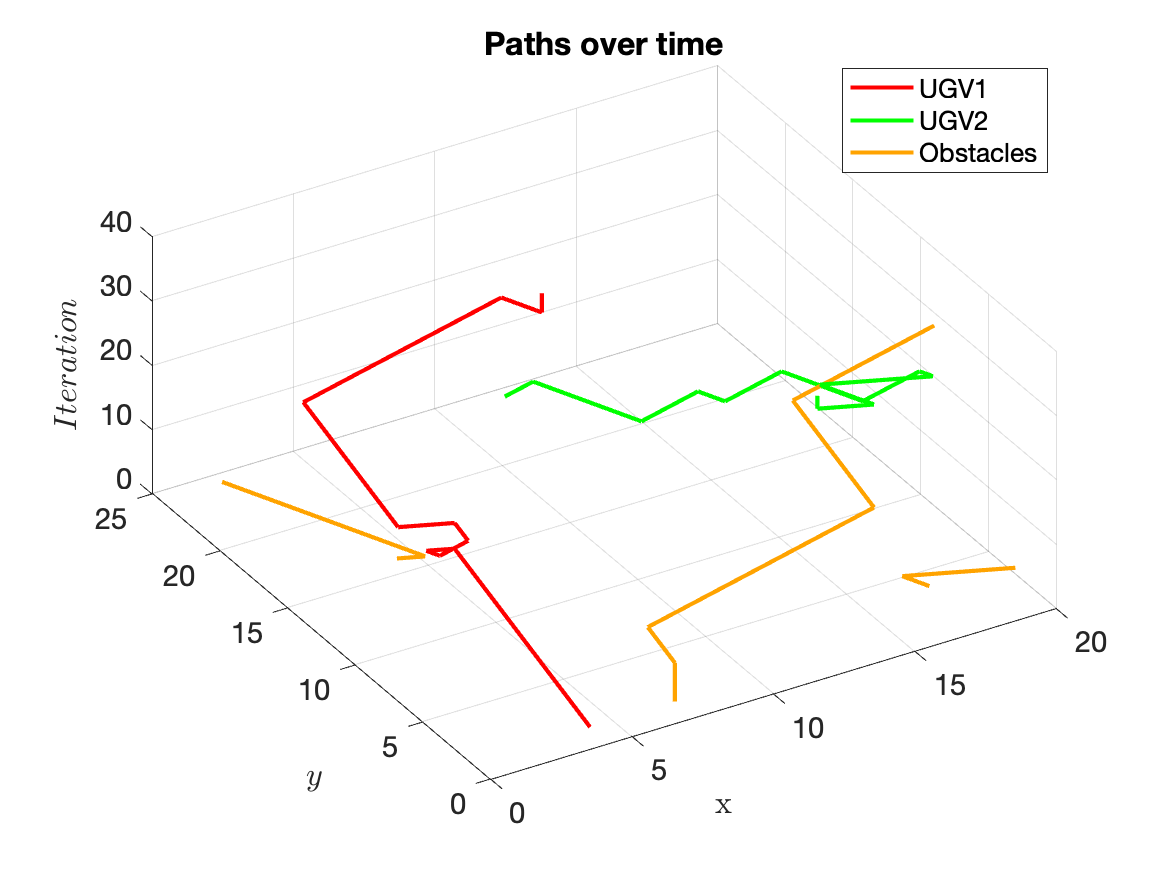} % Replace with your file
        \caption{Regular dynamic obstacle}
        \label{fig:regular_3D}
    \end{subfigure}
    % Add horizontal space
    % \hspace{0.05\linewidth}
    % Second subfigure
    \begin{subfigure}[t]{0.6\linewidth} % Adjust width as needed
        \centering
        \includegraphics[width=\linewidth]{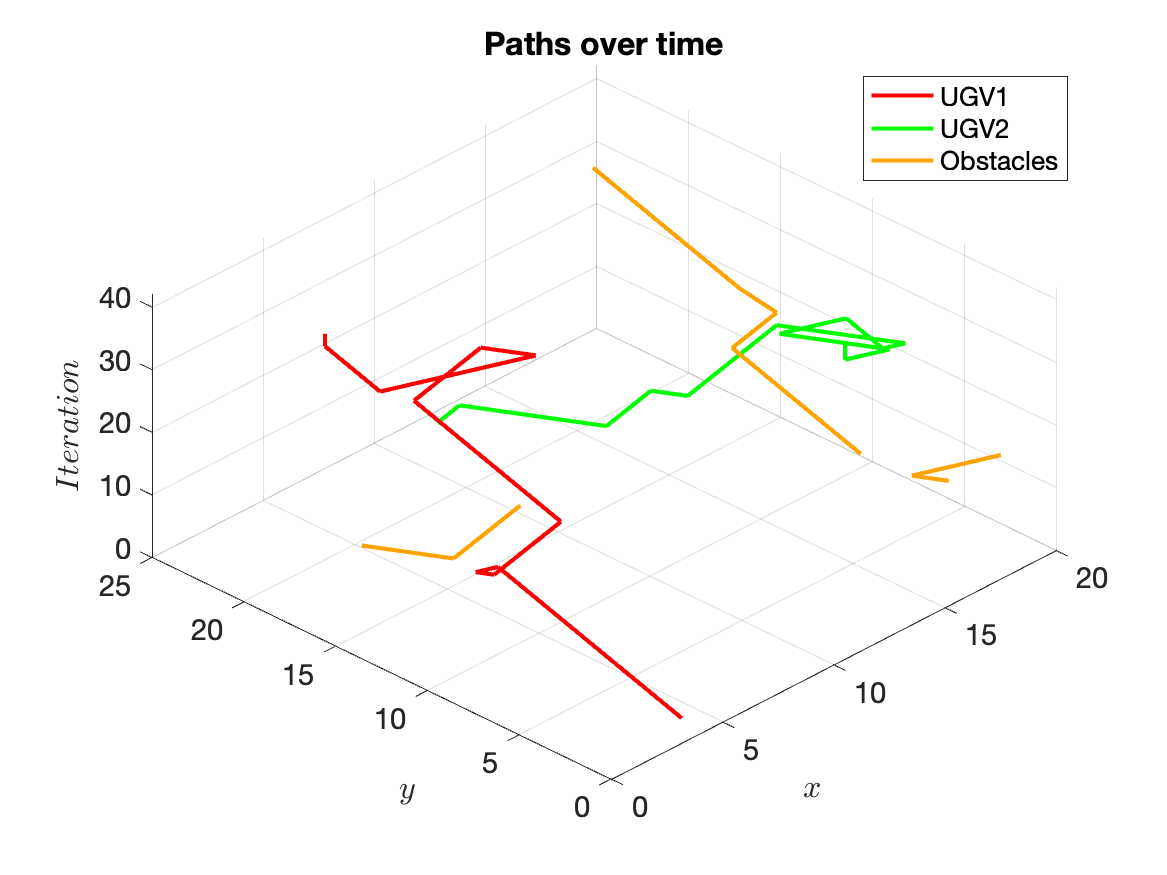} % Replace with your file
        \caption{Sudden dynamic obstacle}
        \label{fig:sudden_3D}
    \end{subfigure}
    \caption{Visualization of UGVs and obstacle trajectories over iterations, demonstrating successful dynamic obstacle avoidance. (a) Regular obstacle movement scenario. (b) Sudden obstacle appearance scenario.}
    \label{fig:3D_view}
\end{figure}

Figure \ref{fig:obs1_b} illustrates UGV paths with obstacles, emphasizing the necessity of adaptive planning, while Fig. \ref{fig:with_obstacle_cost} presents the corresponding cost analysis, demonstrating the impact of obstacle-induced deviations on path efficiency. The initial cost is higher than in the baseline (no-obstacle) scenario, since the early presence of obstacles increases the corresponding edge weights. Also, for clarity, Figure \ref{fig:regular_3D} overlays the UGV and obstacle paths over time, confirming collision-free operation.

\begin{figure}
    \centering
    \includegraphics[width=.20 \textwidth]{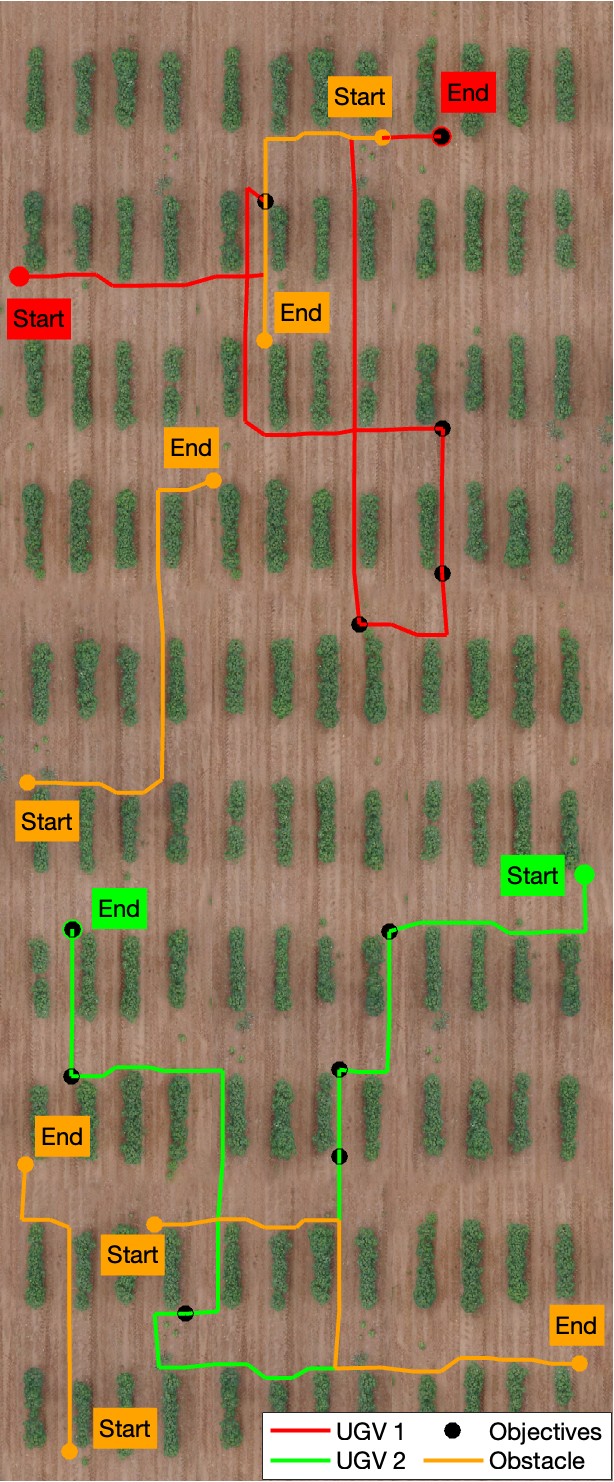}
    \caption{UGV trajectory adjustments in response to a sudden obstacle detected in the field, showcasing real-time re-planning capability. See Fig. \ref{fig:sudden_3D} for the overlap between the UGV and obstacle trajectories.} 
    \label{fig:sudden_obs}
\end{figure}

\begin{figure}
    \centering
    \includegraphics[width=.24 \textwidth]{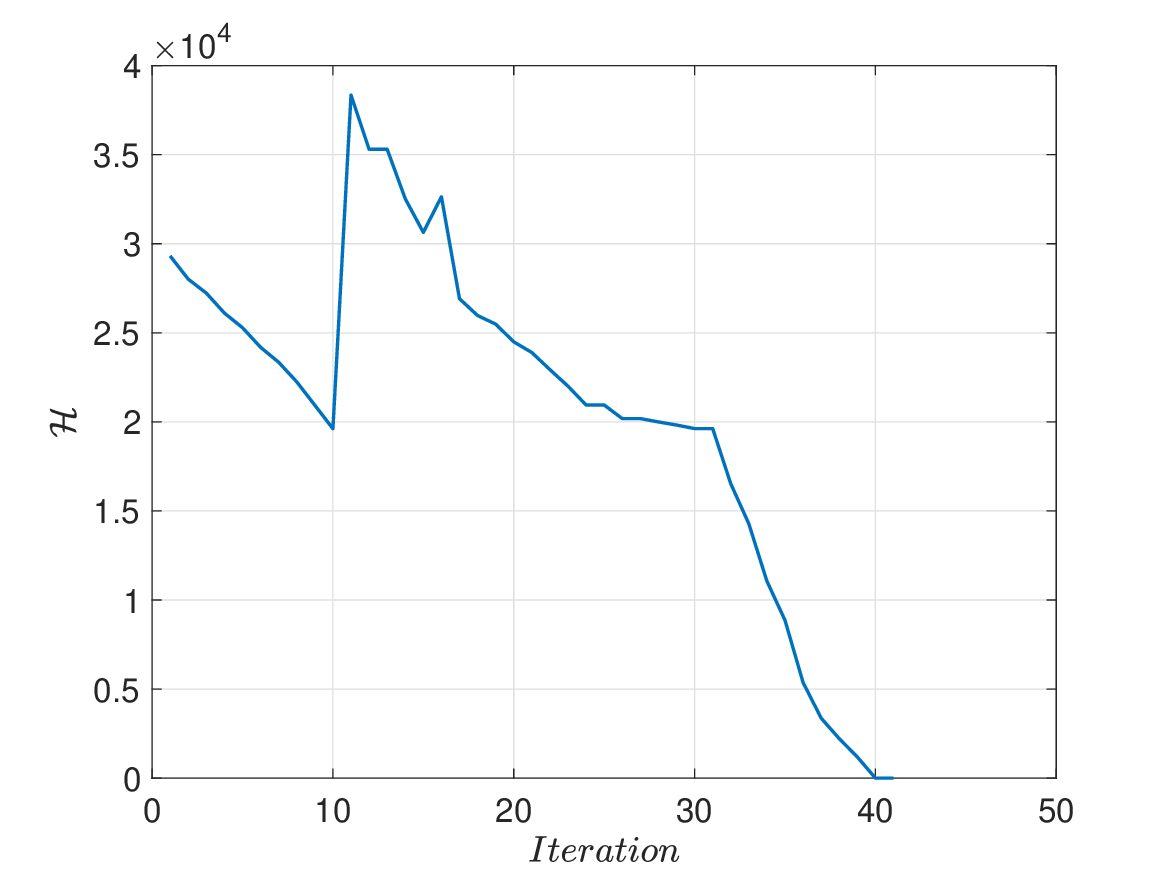}
    \caption{Coverage cost corresponding to sudden obstacles (Fig. \ref{fig:sudden_obs}), demonstrating the impact of sudden obstacles on path efficiency. Note that the cost spikes at obstacle detection instants (iterations 10 and 16) and gradually decreases as adaptive navigation adjusts the path.}
    \label{fig:sudden_obs_cost}
\end{figure}

Now, in a more complicated scenario, if the UAV fails to detect an obstacle initially and a dynamic obstacle appears unexpectedly in the UGV's planned trajectory, the system must promptly find an alternative path to adapt to environmental changes. This rapid adjustment helps prevent potential collisions and operational disruptions. Such a capability is crucial for ensuring uninterrupted and reliable field coverage, particularly in dynamic agricultural environments (see Fig. \ref{fig:sudden_obs}). Also, as shown in Fig. \ref{fig:sudden_obs_cost}, when obstacles are detected in $Iteration=10$ and $Iteration=16$, the coverage cost spikes but subsequently reduces.  Figure \ref{fig:sudden_3D} overlays the UGV and obstacle trajectories over time, confirming collision-free operation. These results emphasize the robustness of the proposed approach in handling both predictable and unexpected changes in the environment.

% A more detailed visualization of UGVs and obstacle trajectories over iteration is provided in Fig. \ref{fig:3D_view}. In Fig. \ref{fig:regular_3D}, the UGVs successfully navigate around a regular dynamic obstacle, whereas Fig. \ref{fig:sudden_3D} highlights the re-planning required when a sudden obstacle is detected. These results emphasize the robustness of the proposed approach in handling both predictable and unexpected changes in the environment.

% The path planning for the Unmanned Ground Vehicle (UGV) is designed to dynamically adapt to changing conditions in the field by incorporating real-time data from the UAV. The UAV continuously monitors the field and identifies moving obstacles, providing their positions and velocities to the UGV. This information is integrated into the UGV’s path planning algorithm, which uses a graph-based approach to calculate the optimal route. As new obstacle data is received, the UGV recalculates its path to avoid potential collisions while maintaining efficiency. This dynamic obstacle avoidance mechanism ensures the UGV can navigate safely and effectively in a dynamic agricultural environment, responding promptly to any changes detected by the UAV. Figure \ref{fig:obs1} compares UGV paths in the presence and absence of obstacles, highlighting the need for adaptive path planning. The cost impact of these obstacles on UGV navigation is analyzed in Fig. \ref{fig:obs1_cost}.

\begin{figure}
    \centering
    \includegraphics[width=.25 \textwidth]{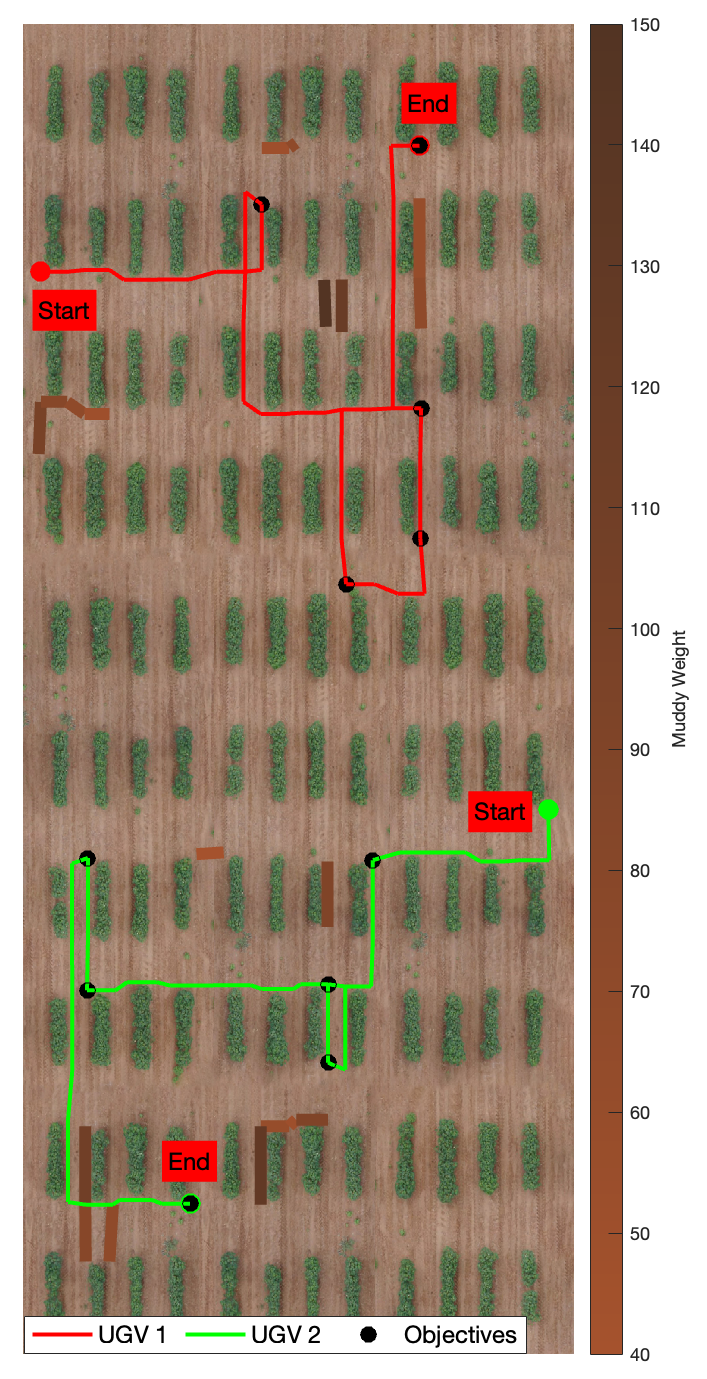}
    \caption{UGVs navigation in an environment with muddy soil patches, where terrain conditions influence path planning decisions. To avoid the muddy areas, UGVs needed to choose a longer path.}
    \label{fig:muddy_2}
\end{figure}

\begin{figure}
    \centering
    \includegraphics[width=.27 \textwidth]{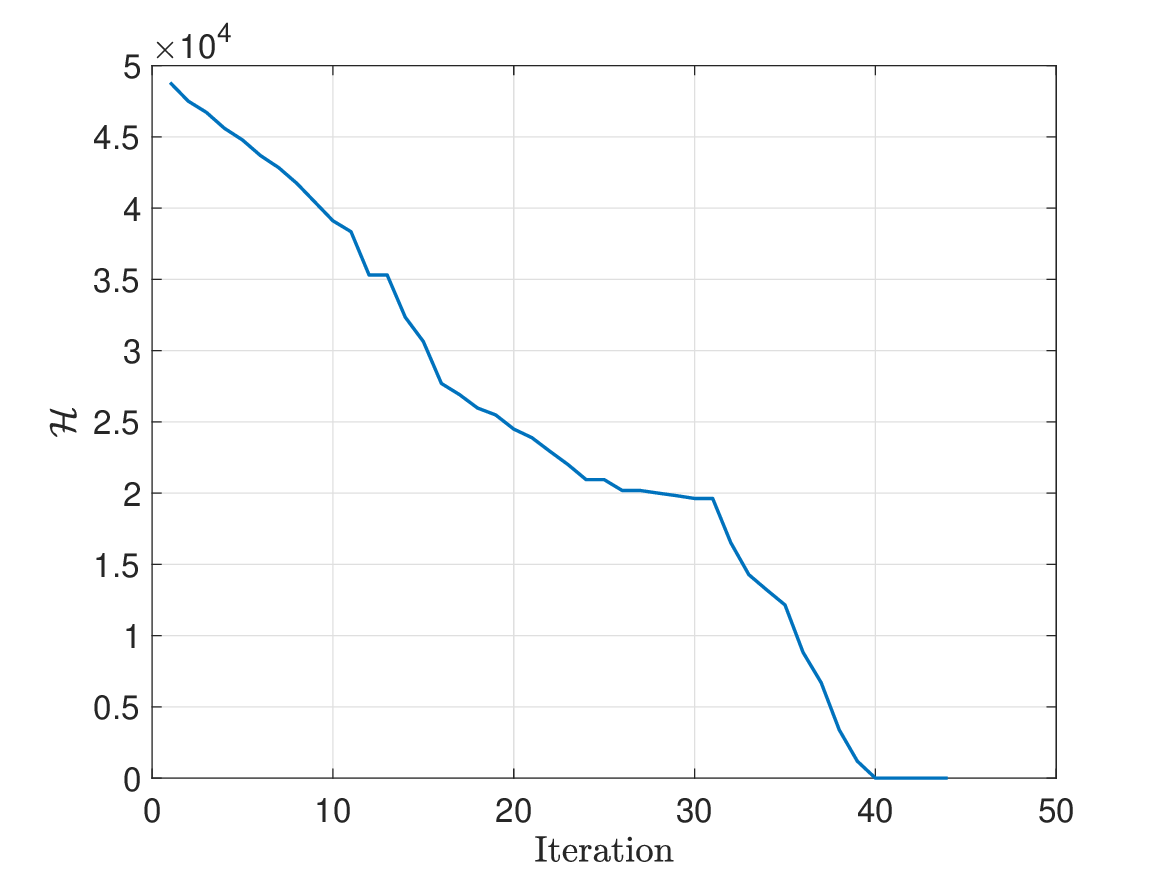}
    \caption{Coverage cost for muddy scenario (Fig. \ref{fig:muddy_2}). As observed, the cost (and paths' length) is increased when compared to the simple (no-obstacle) scenario in Fig. 2(a).}
    \label{fig:muddy_2_cost}
\end{figure}
% \subsection{Dynamic Re-planning for Sudden Obstacles}
% The system’s ability to dynamically re-plan in response to sudden obstacles is a critical component of its robustness. When an unexpected obstacle is detected by the drone, the updated obstacle information is immediately incorporated into the UGV path planning algorithm. This triggers a re-planning process, where the UGV recalculates an alternative path that bypasses the newly identified obstacle while still aiming to minimize travel time and maintain efficiency(Fig \ref{fig:sudden_obs}, \ref{fig:sudden_obs_cost}). The dynamic re-planning mechanism ensures that the UGV can quickly adapt to changes in the environment, avoiding potential collisions and disruptions in its operation. This capability is essential for maintaining continuous and reliable field coverage, especially in unpredictable agricultural settings.
\vspace{-2mm}
\subsection{UGV Path Planning for Muddy Areas}
\vspace{-3mm}
To improve navigation in muddy areas, the UGV path planning employs a weighted graph approach that adjusts for varying terrain conditions. Muddy regions, which hinder UGVs movement, are identified through image analysis or pre-existing field data. These regions are represented as higher-weighted edges in the graph, reflecting increased traversal costs (Figure \ref{fig:muddy_2}). The algorithm prioritizes paths with lower cumulative weights, steering UGVs away from muddy regions whenever possible. This strategy enhances navigation efficiency and reliability by reducing the risk of immobilization and optimizing route selection for smooth field traversal. The impact of muddy terrain on cost function is showed in Figure \ref{fig:muddy_2_cost}. Cost and trajectory can be compared to Figure \ref{fig:obs1_a} and Figure \ref{fig:no_obstacle_cost} when there is no obstacle and muddy region in the field.

\vspace{-2mm}
\section{Conclusion and Future Work} \label{sec:5}
\vspace{-3mm}
% In this paper, we introduced a coverage control strategy for autonomous off-road vehicles operating in dynamic agricultural environments. Our approach combines UAV-based (remote) sensing with UGV path planning in a weighted graph framework, enabling real-time obstacle avoidance and efficient terrain-aware navigation. By leveraging Voronoi-based node assignment, adaptive edge weight updates, and cost-based optimization, our method ensures robust coverage in dynamic environments. Extensive simulation results were provided to demonstrate the effectiveness of the proposed approach.

% Future research aims to extend this framework to incorporate machine learning for predictive obstacle modeling and conduct real-world experiments to validate its use in agriculture, enhancing its adaptability and scalability for precision farming.

In this paper, we proposed a coverage control method for autonomous field navigation that dynamically adapts to obstacles and environmental changes. Our approach combines UAV-based (remote) sensing with UGV path planning in a weighted graph framework, enabling real-time obstacle avoidance and efficient terrain-aware navigation. By leveraging Voronoi-based node assignment, adaptive edge weight updates, and cost-based optimization, our method ensures robust coverage in dynamic environments. Simulation results demonstrated the effectiveness of the proposed approach in handling moving obstacles, adapting to muddy terrains, and re-planning paths efficiently in response to sudden environmental changes. The results confirmed that our strategy significantly improves path planning efficiency while minimizing traversal cost and unnecessary detours.

Future research includes extending the framework to multi-UGV coordination with decentralized decision-making, incorporating machine learning for predictive obstacle modeling, and conducting real-world field experiments to validate the system's performance in practical agricultural scenarios. By further enhancing adaptability and scalability, this approach can contribute to more efficient and autonomous precision agriculture solutions.
\vspace{-2mm}
\bibliography{Ref}

\end{document}